\documentclass[conference]{IEEEtran}
\IEEEoverridecommandlockouts
\usepackage{cite}
\usepackage{amsmath,amssymb,amsfonts}
\usepackage{algorithmic}
\usepackage{graphicx}
\usepackage{textcomp}
\usepackage{booktabs}
\usepackage{multirow}

\newcommand{\tightcmidrule}[1]{%
  \cmidrule(lr){#1}%
}

\usepackage[table]{xcolor}
\usepackage[table]{xcolor}
\definecolor{TCrow}{RGB}{255,245,200} 
\definecolor{BlockA}{RGB}{255,242,204} 
\definecolor{BlockB}{RGB}{189,215,238} 
\definecolor{BlockC}{RGB}{197,224,180} 
\def\BibTeX{{\rm B\kern-.05em{\sc i\kern-.025em b}\kern-.08em
    T\kern-.1667em\lower.7ex\hbox{E}\kern-.125emX}}

\newcommand{\legendbox}[1]{%
  \begingroup
  \setlength{\fboxsep}{0pt}
  \setlength{\fboxrule}{0.1pt}
  \fcolorbox{black}{#1}{\textcolor{#1}{\%}}%
  \endgroup
}
\usepackage{array}
\newcolumntype{C}[1]{>{\centering\arraybackslash}p{#1}}
\usepackage[T1]{fontenc}

\usepackage{color}
\usepackage{url}

\usepackage{hyperref}
    
\begin{document}

\title{Fed-ADApt: Federated Anytime Depth Adaptation for Resource-Aware Medical Image Segmentation
}

\author{
\IEEEauthorblockN{Abhijeet Parida}
\IEEEauthorblockA{
\textit{Sheikh Zayed Institute for Pediatric Surgical Innovation,}\\
\textit{Children's National Hospital}, Washington, DC, United States\\
\textit{Escuela T\'ecnica Superior de Ingenieros de Telecomunicaci\'on,}\\
\textit{Universidad Polit\'ecnica de Madrid}, Madrid, Spain\\ \texttt{pabhijeet@childrensnational.org}}
\and
\IEEEauthorblockN{Zhifan Jiang}
\IEEEauthorblockA{
\textit{Sheikh Zayed Institute for Pediatric Surgical Innovation,}\\
\textit{Children's National Hospital}, Washington, DC, United States}
\and
\IEEEauthorblockN{Pooneh Roshanitabrizi}
\IEEEauthorblockA{
\textit{Sheikh Zayed Institute for Pediatric Surgical Innovation,}\\
\textit{Children's National Hospital}, Washington, DC, United States}
\and
\IEEEauthorblockN{Austin Tapp}
\IEEEauthorblockA{
\textit{Sheikh Zayed Institute for Pediatric Surgical Innovation,}\\
\textit{Children's National Hospital}, Washington, DC, United States}
\and
\IEEEauthorblockN{Mar\'ia J. Ledesma-Carbayo}
\IEEEauthorblockA{
\textit{Escuela T\'ecnica Superior de Ingenieros de Telecomunicaci\'on,}\\
\textit{Universidad Polit\'ecnica de Madrid}, Madrid, Spain\\
\textit{CIBER-BBN, Instituto de Salud Carlos III}, Madrid, Spain}
\and
\IEEEauthorblockN{Syed Muhammad Anwar}
\IEEEauthorblockA{
\textit{Sheikh Zayed Institute for Pediatric Surgical Innovation,}\\
\textit{Children's National Hospital}, Washington, DC, United States\\
\textit{Departments of Radiology and Pediatrics,}\\
\textit{The George Washington University}, Washington, DC, United States}
\and
\IEEEauthorblockN{Ziyue Xu}
\IEEEauthorblockA{
\textit{NVIDIA Corporation}, Santa Clara, CA, United States}
\and
\IEEEauthorblockN{Marius George Linguraru}
\IEEEauthorblockA{
\textit{Sheikh Zayed Institute for Pediatric Surgical Innovation,}\\
\textit{Children's National Hospital}, Washington, DC, United States\\
\textit{Departments of Radiology and Pediatrics,}\\
\textit{The George Washington University}, Washington, DC, United States}
\and
\IEEEauthorblockN{Holger R. Roth}
\IEEEauthorblockA{
\textit{NVIDIA Corporation}, Santa Clara, CA, United States}
}

\maketitle

\begin{abstract}
Federated learning (FL) enables collaborative training of medical image segmentation models without sharing raw patient data, yet existing approaches assume a homogeneous compute budget across institutions, limiting participation of low-resource sites. We propose \textit{Fed-ADApt}, a depth-adaptive federated framework for \textit{UNet}-based segmentation that jointly addresses low-compute training and inference. \textit{Fed-ADApt} integrates multi-depth supervision with hierarchical depth-wise aggregation, allowing each site to train according to its local compute budget while contributing to a global model that supports dynamic depth selection at deployment. We evaluated \textit{Fed-ADApt} on multi-site 2D retinal fundus disc segmentation and 3D brain tumor segmentation. Across both tasks, federated collaboration substantially improves robustness under domain shift. \textit{Fed-ADApt} matched the full-resource \textit{FedAvg} performance in 3D and achieved competitive 2D performance with a 4.7\% average Dice reduction, while reducing average inference cost by 19.5\% in 3D and 34.5\% in 2D and substantially reducing training cost by 98\% at the most constrained sites. Importantly, \textit{Fed-ADApt} enables low-resource institutions that cannot train full-capacity models to participate in federations while maintaining competitive global performance under a favorable accuracy to efficiency trade-off. By considering training and inference compute budgets, \textit{Fed-ADApt} provides a practical and equitable solution for federated medical image segmentation across heterogeneous clinical and edge-enabled imaging environments.
\end{abstract}


\section{Introduction}

Federated learning (FL) enables collaborative training of machine learning models across institutions without sharing data~\cite{diao2021heterofl,GUAN2024110424}. \textit{UNet}-like architectures~\cite{unet} are standard for medical image segmentation and widely adopted in FL~\cite{qi2022contrastive,10004567,3235757}. Beyond large hospital-based imaging systems, medical imaging is increasingly being extended through connected and point-of-care devices, including portable magnetic resonance imaging (MRI) systems, handheld fundus cameras, and edge-connected clinical workstations~\cite{kagita2022review,deoni2022development,chen2025acceptability}. These Internet of Medical Things (IoMT) devices offer a promising lower-cost pathway to expand access to diagnostic imaging in low-resource settings, where conventional imaging infrastructure and high-performance computing resources may not be available. However, such devices often operate under strict memory, compute, bandwidth, power, and latency constraints~\cite{kaur2023federated}. Therefore, a deployable federated medical imaging system must not only preserve patient privacy across sites, but also support clients whose local hardware cannot train or deploy a full-capacity segmentation model.
\begin{figure*}[t!]
\centering
\includegraphics[width=0.8\textwidth]{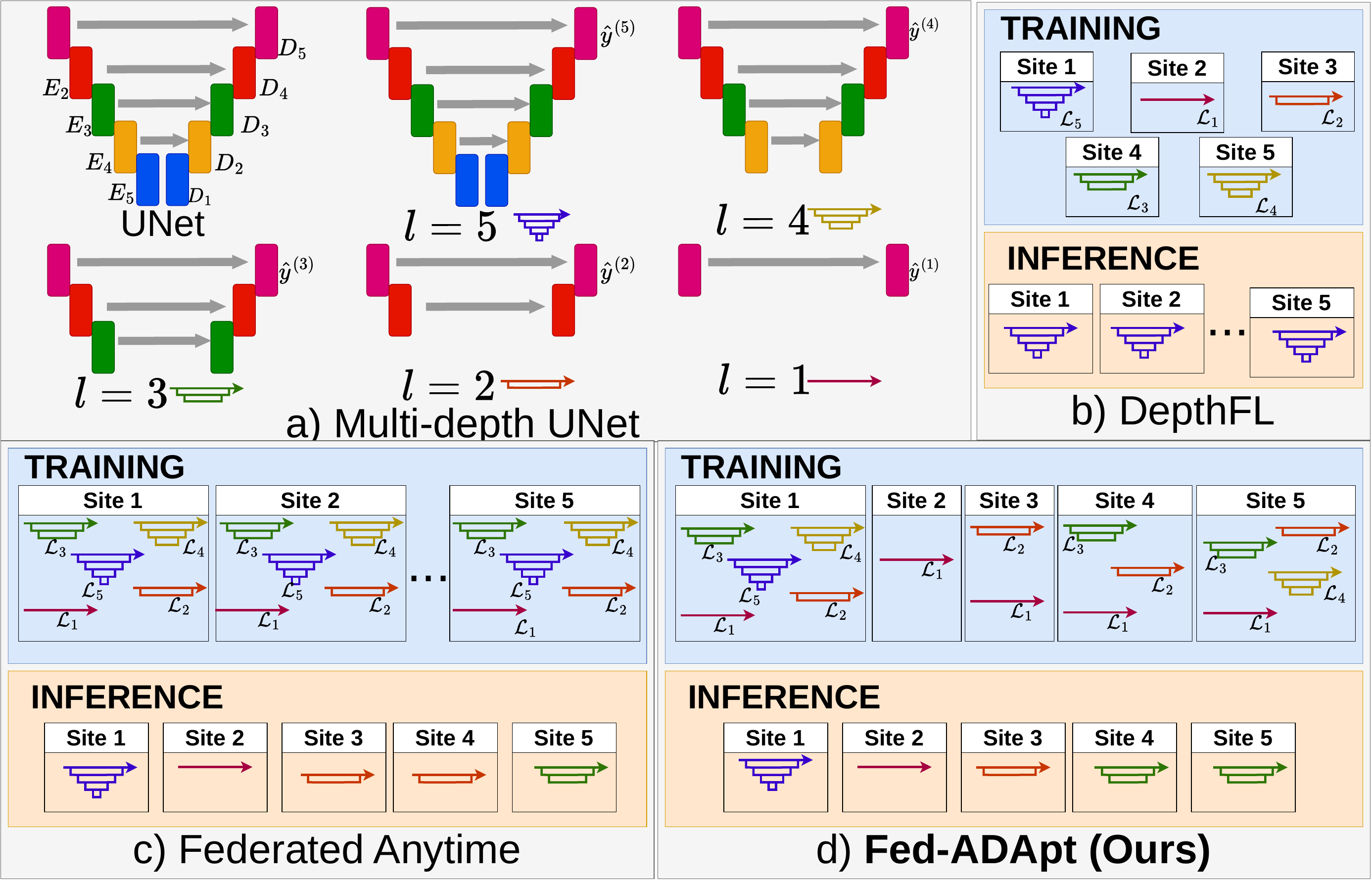}
\caption{\textbf{Fed-ADApt for resource-aware federated segmentation.}
(a) The proposed multi-depth UNet produces full-resolution predictions from multiple encoder depths, enabling different compute paths within a single architecture.
(b) \textit{DepthFL} supports heterogeneous training by allowing clients to train depth-restricted models, but all clients perform full-depth inference.
(c) \textit{Federated anytime} supports adaptive inference by training all depths at every client, but requires full-model training at all sites.
(d) \textbf{Fed-ADApt} jointly addresses both constraints by allowing each client to train only the depths permitted by its compute budget while retaining dynamic depth selection at inference.}\label{anyexit_unet}
\end{figure*}
However, existing FL approaches assume that all participating sites train and deploy identical models with similar compute resources~\cite{diao2021heterofl,10.1145/3625558,zhang2023memory}. This assumption has hindered the adoption of FL in low-resource sites, especially in developing countries.
In practice, even though institutions participating in FL share the same task (e.g., brain tumor segmentation), they may differ in hardware capabilities and latency requirements~\cite{10825769,10.1145/3625558}. Some institutions face strict constraints on training cost, memory capacity, or inference latency~\cite{zhang2023memory}. These constraints are amplified in IoMT environments, where cheaper imaging devices increase data availability,  but may not have the compute capacity of hospital servers. This creates a mismatch between the clinical value of distributed imaging data and the computational requirements of standard FL segmentation models. We address medical image segmentation under varying training and deployment computing budgets for FL to (i) enable low-resource sites to participate in training and (ii) support compute-adaptive inference at deployment time, while maintaining the performance of the FL model.

Standard FL algorithms such as \textit{FedAvg}~\cite{fedavg} assume architectural uniformity across sites. FL knowledge distillation methods, such as \textit{FedAgent}~\cite{SUN2025127973} or \textit{FedKD}~\cite{wu2022communication}, transfer model-agnostic knowledge by averaging site logits on shared or synthetic data. Their reliance on additional data at the server and on teacher–-student alignment in dense predictions for 2D or 3D segmentation can lead to inconsistencies. Recent heterogeneous FL allows sub-network training~\cite{diao2021heterofl,10825769,litany2022federated}. However, these approaches target image classification models that follow simple sequential architectures. Unlike classification models, segmentation models, such as \textit{UNet}, have encoder–-decoder architectures with skip connections that transfer features at multiple depth levels. Since segmentation requires multi-scale feature integration for pixel-level prediction, naive sub-network extraction can degrade performance, making heterogeneous FL in \textit{UNet} non-trivial.

Existing heterogeneous FL methods predominantly address training-time heterogeneity, allowing sites to update only subsets of parameters under limited compute budgets~\cite{diao2021heterofl,kim2023depthfl}. \textit{HeteroFL}~\cite{diao2021heterofl} addresses the train compute heterogeneity by varying channel widths at each network level. This may be less effective for capturing hierarchical representations~\cite{kim2023depthfl}. In contrast, \textit{DepthFL}~\cite{kim2023depthfl} addresses compute heterogeneity by allowing sites to train models of varying depths. In parallel, anytime and early-exit models focus on inference-time heterogeneity, enabling dynamic trade-offs between inference compute cost and accuracy during deployment~\cite{lee2024recurrent,liu2022federated}. These two forms of heterogeneity are typically treated in isolation. However, a deployable clinical FL system must accommodate both: institutions lacking resources to train full-capacity models and constrained deployment conditions. To our knowledge, no unified framework that jointly addresses training and inference heterogeneity for \textit{UNet}-based medical image segmentation exists. 

This setting is different from previous heterogeneous FL methods for classification architectures, where subnetworks can often be defined by truncating sequential convolutional or transformer backbones. In UNet-based segmentation, depth heterogeneity must preserve encoder--decoder compatibility, skip-connected feature transfer, and full-resolution dense prediction. These architectural constraints make depth-adaptive FL for segmentation non-trivial and require a design that jointly supports hierarchical aggregation during training and dynamic depth selection during inference.

To bridge this gap, we propose \textit{Fed-ADApt}, a depth-adaptive framework for federated segmentation that supports adaptive training, lower communication costs, and anytime inference at deployment. Our approach adds multi-depth supervision to \textit{UNet} architectures, enabling hierarchical depth-wise aggregation in FL. Each site trains according to its local compute capacity while contributing to a unified global model that allows dynamic depth selection during deployment for inference. 
\noindent \textbf{Our contributions} are: 
\begin{enumerate}
\item We generalize depth-wise aggregation of \textit{DepthFL} to encoder–decoder architectures with skip connections, enabling heterogeneous FL training in \textit{UNet}.
\item We integrate multi-depth supervision within FL for \textit{UNet}, enabling compute-adaptive inference while preserving segmentation performance.
\item We introduce \textit{Fed-ADApt}, a budget-aware optimization framework that supports depth-adaptive participation during training, dynamic depth selection at inference, and reduces communication cost.
\item We demonstrate on two real-world multi-site datasets that \textit{Fed-ADApt} matches the Dice performance of \textit{FedAvg} in 3D and provides a favorable accuracy-efficiency trade-off in 2D while reducing inference, communication, and low-resource-client training costs.
\end{enumerate}

\section{Method}

We consider an FL setup with $S$ clinical sites indexed by
$s \in \{1, .., S\}$. Each site $s$ has a dataset
$\mathcal{D}_s = \{(x_i^s, y_i^s)\}_{i=1}^{N_s}$, where $x_i^s$ is an input image
and $y_i^s$ its segmentation mask. The sites differ in compute resources, given by a normalized training budget $C_s^{\mathrm{train}} \in {1,\ldots,5}$, where $C_s^{\mathrm{train}}=5$ indicates full-capacity training. This budget determines the maximum depth a site can instantiate during local optimization, while deployment-time inference may use a separate budget $C_s^{\mathrm{infer}}$ because inference is forward-only and cheaper than training. 

Our goal is to learn an FL model that supports heterogeneous $C_s^{\mathrm{train}}$ and $C_s^{\mathrm{infer}}$ during training and inference while maintaining good performance at all $s$. All sites solve the same segmentation task and
share a common label space. The FL objective is: ~~$
\min_{\theta} \sum_{s=1}^{S} \frac{N_s}{\sum_{j=1}^{S} N_j}
\mathbb{E}_{(x,y)\sim\mathcal{D}_s}
\left[ \mathcal{L}(f_{\theta_s}(x), y) \right],
$
where $\theta_s \subseteq \theta$ are the parameters accessible to
site $s$ with compute budget $C_s^{\mathrm{train}}$.

\subsection{Multi-depth UNet}\label{sec:me_unet}
We consider a \textit{UNet} with five encoder
$\{E_1,..,E_5\}$ and  corresponding decoder 
$\{D_5,..,D_1\}$ blocks. \textit{UNet} performs forward propagation
through all encoder and decoder blocks, assuming uniform
compute budget at sites. We construct a \emph{multi-depth
UNet} by enabling prediction from multiple encoder depths while decoding each selected depth to full resolution. Let 
\begin{equation}\label{eqn:encoder}
z_l = E_l \circ E_{l-1} \circ \cdots \circ E_1(x), \quad l \in \{1,..,5\},
\end{equation}
be the representation after block $E_l$. For each depth $l$, we define a depth-specific prediction path that decodes $z_l$ using the corresponding decoder:
$\hat{y}^{(l)} = \Phi^{(l)}(z_1,..,z_l)$,
where $\Phi^{(l)}$ consists of decoder blocks
$\{D_{6-l},..,D_5\}$ with skip connections from the available encoder features. Concisely, as shown in  Fig.~\ref{anyexit_unet}a, for all $l \in {1,..,5}$, \  $\hat{y}^{(l)}$ uses the 
decoder blocks required to upsample the selected encoder representation to full resolution, by $D_5$, with skip connections from the available encoder features.
The depth $l=5$ corresponds to the standard full \textit{UNet}, while shallower depths allow for reduced computation.

Unlike conventional deep supervision, where auxiliary heads primarily provide training losses at intermediate resolutions, each depth-specific path in the proposed multi-depth \textit{UNet} produces a full-resolution segmentation output and can be independently selected at inference. Thus, these predictions are not merely auxiliary optimization signals; they define compute-adaptive inference paths under different resource budgets.

\subsection{DepthFL UNet}
\label{sec:depthfl}
\begin{table*}[ht]
\centering
\caption{\textbf{Multi-site datasets and simulated compute budgets.}
Summary of the 2D retinal fundus and 3D brain tumor segmentation datasets used for federated evaluation. Each site is assigned a normalized compute budget $C_s^{\mathrm{train}} \in {1,\ldots,5}$, where lower values represent resource-constrained clients and $C_s^{\mathrm{train}}=5$ denotes full-capacity training.}
\label{tab:dataset}
\resizebox{0.85\textwidth}{!}{%
\begin{tabular}{l c c c c c || C{1.1cm}C{1.1cm}C{1.1cm}C{1.1cm}C{1.1cm}}
\toprule
& \multicolumn{5}{c||}{\textbf{Retina Disc Segmentation (2D)}} 
& \multicolumn{5}{c}{\textbf{Brain Tumor Segmentation (3D)}~\cite{brats}} \\

& Site A 
& Site B 
& Site C 
& Site D 
& Site E 
& Site P 
& Site Q  
& Site R 
& Site S  
& Site T \\

& Drishti\cite{drishti}
& BinRushed4\cite{riga}
& Magrabia\cite{riga} 
& Messidor\cite{riga}
& REFUGE\cite{refuge} 
&UCSF 
& Duke
& UCSD 
& Missouri
& Indiana \\
\midrule
$C_s^{\mathrm{train}}$ 
& 1 & 2 & 1 & 4 & 5 
& 5 & 4 & 3 & 3 & 1 \\
\tightcmidrule{1-11}
Train 
& 50 & 98 & 47 & 230 & 400 
& 357 & 284 & 120 & 192 & 18 \\

Val 
& 25 & 49 & 24 & 115 & 200 
& 119 & 95 & 40 & 64 & 6 \\

Test 
& 25 & 49 & 23 & 115 & 200 
& 120 & 95 & 40 & 64 & 7 \\

\bottomrule
\end{tabular}
}
\end{table*}
To support heterogeneous training under constrained compute resources, we
adopt a depth-adaptive federated learning strategy for \textit{UNet}, inspired by
\textit{DepthFL}~\cite{kim2023depthfl}. In contrast to multi-depth supervision,
each site trains a \emph{single-depth UNet} whose depth is determined by $C_s^{\mathrm{train}}$.

\noindent\textbf{Training:}
At $s$, $C_s^{\mathrm{train}}$ specifies the maximum encoder depth that can
be instantiated (Fig.~\ref{anyexit_unet}b). The local model consists of encoder blocks
$\{E_1,..,E_{C_s}\}$ and the corresponding decoder blocks
$\{D_{6-C_s},..,D_5\}$, which produce a single prediction
$\hat{y}^{(C_s^{\mathrm{train}})}$ from depth $l=C_s^{\mathrm{train}}$. Predictions from depths other than $l$ are  not available in $s$. Each site optimizes segmentation loss, $\mathcal{L}$, at its local depth:
\begin{equation}
\mathcal{L}_{\text{depth}}(x,y)
=
\mathcal{L}(\hat{y}^{(C_s^{\mathrm{train}})}, y).
\end{equation}
Thus, supervision is applied exclusively at the maximum depth $l$, supported by $s$, resulting in a single loss training objective.
Let $\theta^{(l)}$ denote the encoder and decoder parameters associated with depth $l$. During FL, updates to $\theta^{(l)}$ are produced only by sites whose compute budgets satisfy $C_s^{\mathrm{train}} \ge l$. In particular, depth $l+1$ contains all parameters of depth $l$ plus the additional encoder block $E_{l+1}$ and its corresponding decoder components such that $\theta^{(l)} \subset \theta^{(l+1)}$. On the server, the parameters are aggregated
independently by depth. For each $l \in \{1,..,5\}$, define the set of participating sites $\mathcal{S}_l = \{ s \mid C_s^{\mathrm{train}} \ge l \}$.
The server update for all depths $l$ given by
\begin{equation}\label{eqn:agg}
\theta^{(l)} \leftarrow
\sum_{s\in\mathcal{S}_l}
\frac{N_s}{\sum_{j\in\mathcal{S}_l} N_j}\,\theta^{(l)}_s.
\end{equation}
As a result, shallow layers receive updates from all sites, while deeper layers are trained only by high $C_s^{\mathrm{train}}$ sites.

\noindent\textbf{Inference:}
After training,  as seen in Fig.~\ref{anyexit_unet}b, each site $s$ performs inference using the global model of maximum depth($l=5$), independent of its deployment-time inference budget $C_s^{\mathrm{infer}}$. Thus, \textit{DepthFL} addresses heterogeneous training capacity but does not provide compute-adaptive inference.

\subsection{Federated Anytime UNet}
\label{sec:fed_anytime}

Inspired by training multiple exits in a linear model in \cite{lee2024recurrent,liu2022federated}, we propose
\emph{federated anytime UNet}, which jointly optimizes multiple depths from our \emph{multi-depth UNet} (Sec.~\ref{sec:me_unet}) during
FL and enables flexible and compute-adaptive inference without retraining.

\noindent\textbf{Training:}
Each site instantiates a full \emph{multi-depth UNet} with prediction paths
at all encoder depths $l \in \{1,..,5\}$ as in Eqn.~\ref{eqn:encoder}. For $(x,y)$, the \textit{UNet}
produces 
$\{\hat{y}^{(1)}, .., \hat{y}^{(5)}\}$, corresponding to all $l$. All outputs are supervised using the loss function:
\begin{equation}
\label{eqn:multiexit}
\mathcal{L}_{\text{any}}(x,y)
=
\sum_{l=1}^{5}
\mathcal{L}(\hat{y}^{(l)}, y),
\end{equation}
This supervision encourages early depths to learn rough predictions, while allowing deeper depths to refine predictions. We use equal weighting to avoid depth bias and empirically found it to be stable. Since all sites train the full model and update all parameters, it is a resource-intensive 
optimization using \textit{FedAvg}~\cite{fedavg}, using $l=5$ in Eqn.~\ref{eqn:agg}.

\noindent\textbf{Inference:}
After training, the server maintains a single global model that is optimized across all depths. At inference time, as seen in Fig.~\ref{anyexit_unet}c, each site can dynamically select a depth
$l$ based on its available compute budget $C_s^{\mathrm{infer}}$, such that $l \le C_s^{\mathrm{infer}}$ and achieves the highest validation score. Therefore, by selecting $l \le C_s^{\mathrm{infer}}$, each site $s$ obtains the best possible segmentation accuracy while staying within the compute budget without modifying or retraining the model.

\subsection{Federated Anytime
Depth Adaptation  (Fed-ADApt)}
\label{sec:adapt_unet}

\noindent
While \textit{DepthFL} and \textit{federated anytime UNet} enable heterogeneous training  and flexible inference, respectively, neither provides a solution
for clients where both training and inference are constrained. To overcome this, we propose \textit{Fed-ADApt} (Fig.~\ref{anyexit_unet}d).

\noindent \textbf{Training:} The compute budget $C_s^{\mathrm{train}} = l$ determines the maximum network depth that can be instantiated at $s$. Unlike \textit{federated anytime UNet}, where all sites train every depth, \textit{Fed-ADApt} similar to \textit{DepthFL}, restricts the network depth and  hence the number of supervised depths to keep under the budget $C_s^{\mathrm{train}}$. Specifically, site
$s$ instantiates encoder blocks $\{E_1,..,E_l\}$, from Eqn.~\ref{eqn:encoder}, decoder blocks $\{D_{6-l},..,D_5\}$, and depth-specific predictions $\{\hat{y}^{(1)},..,\hat{y}^{(l)}\}$.
For an input–label pair $(x, y)$, the site jointly supervises all available depths
up to depth $l$ using multi-depth loss in Eqn. \ref{eqn:multiexit}.
This design allows shallow depths to receive supervision while avoiding the computational overhead of deeper layers that are infeasible at $s$. Parameter aggregation follows \textit{DepthFL} using Eqn.~\ref{eqn:agg} where sites with $C_s^{\mathrm{train}} \ge l$ contribute the corresponding global $\theta^{(l)}$. This enables generalizable knowledge transfer from resource rich to low-resource sites at the shallow layers, while deeper layers are specialized using updates from high budget
institutions.

\noindent\textbf{Inference:}
After training, the server maintains a single global \textit{Fed-ADApt} model with prediction paths at multiple depths. At inference, each site dynamically selects a depth $l_s^{\mathrm{infer}}$ that achieves the best validation score within its deployment-time budget $C_s^{\mathrm{infer}}$. Since inference does not require backpropagation or optimizer states, $C_s^{\mathrm{infer}}$ may allow a deeper prediction path than the training budget $C_s^{\mathrm{train}}$.
\section{Experiments and Results}

\paragraph{\textbf{Dataset}}
\textit{Fed-ADApt} was evaluated on two multi-site medical segmentation tasks covering both 2D and 3D settings (Table~\ref{tab:dataset}). To simulate realistic compute heterogeneity, each site was assigned a training budget $C_s^{\mathrm{train}} \in \{1,..,5\}$ proportional to its data, reflecting the assumption that larger clinics typically have greater compute resources. Because training involves forward and backward propagation and weight updating, it is three times more computationally expensive than inference\cite{cost}, the deployment-time inference budget was defined as $C_s^{\mathrm{infer}} = min(5, \  3 \times C_s^{\mathrm{train}})$. This allows for sites to use deeper prediction paths during inference than during local training when computationally feasible.
To assess robustness to the assumed budget assignment, we also evaluate ``low'', ``medium'', and ``high'' federation budget settings while keeping the lowest and highest resource sites fixed.

\begin{table*}[t]

\caption{\textbf{2D retinal disc segmentation.} Dice scores and compute cost for various FL methods under compute constraints. GMAC was calculated for a batch size of 10. $^*$ indicates statistical significance ($p < 0.05$) in Dice compared to \textit{Fed-ADApt}. Training strategy: \legendbox{BlockA} No FL; 
\legendbox{BlockB} full-capacity FL; \legendbox{BlockC} compute-aware FL.}

\centering

\resizebox{\textwidth}{!}{%
\begin{tabular}{lcccccc||ccc|ccc|c}
\toprule
& \multicolumn{6}{c||}{\textbf{Segmentation}}
& \multicolumn{3}{c|}{\textbf{Training}}
& \multicolumn{3}{c|}{\textbf{Inference}}& \textbf{Comm.} \\
\textbf{Method} 
& \multicolumn{6}{c||}{\textbf{(Dice)}}
& \multicolumn{3}{c|}{\textbf{GMAC / VRAM(GB)}}
& \multicolumn{3}{c|}{\textbf{GMAC / VRAM(GB)}} & \textbf{Cost}\\

& Site A & Site B & Site C & Site D & Site E & Avg
& Max & Min & Avg
& Max & Min & Avg & \\
\midrule

\rowcolor{BlockA}
Local Model
& 0.27$\pm$0.11& 0.70$\pm$0.18& 0.20$\pm$0.08 & 0.80$\pm$0.06& 0.87$\pm$0.13 & 0.55$\pm0.11$$^*$
& 16.5 / 0.04 & 0.7 / 0.03 & 6.6 / 0.03
& 5.5 / 0.04 & 0.2 / 0.02 & 2.2 / 0.02  & -\\
\rowcolor{BlockA}
High Resource Model
& 0.41$\pm$0.26& 0.06$\pm$0.03& 0.21$\pm$0.13& 0.07$\pm$0.04& 0.87$\pm$0.13& 0.32$\pm$0.12$^*$
& 16.5 / 0.04 & 0.0 / 0.00 & 3.3 / 0.01
& 5.5 / 0.04 & 5.5 / 0.04 & 5.5 / 0.04& -\\

\rowcolor{BlockB}
FedAvg
& 0.90$\pm$0.10 & 0.80$\pm$0.17 & 0.91$\pm$0.05 & 0.80$\pm$0.13 & 0.82$\pm$0.14 & 0.85$\pm$0.12$^*$
& 16.5 / 0.04 & 16.5 / 0.04 & 16.5 / 0.04
& 5.5 / 0.04 & 5.5 / 0.04 & 5.5 / 0.04 & 100\%\\
\rowcolor{BlockB}
Federated Anytime
& 0.88$\pm$0.13 & 0.75$\pm$0.15 & 0.88$\pm$0.16 & 0.86$\pm$0.08 & 0.83$\pm$0.15 & 0.84$\pm$0.13$^*$
& 33.9 / 0.12& 33.9 / 0.12 & 33.9 / 0.12
& 5.5 / 0.04 & 2.5 / 0.03 & 3.9 / 0.04& 100\%\\

\rowcolor{BlockC}
DepthFL
& 0.57$\pm$0.18 & 0.17$\pm$0.10 & 0.45$\pm$0.21 & 0.20$\pm$0.09 & 0.82$\pm$0.13 & 0.44$\pm$0.18$^*$
& 16.5 / 0.04 & 0.7 / 0.03 & 6.6 / 0.03
& 5.5 / 0.04 & 5.5 / 0.04 & 5.5 / 0.04 & 40\%\\
\rowcolor{BlockC}
\textbf{Fed-ADApt (Ours)}
& 0.86$\pm$0.15 & 0.68$\pm$0.17 & 0.86$\pm$0.19 & 0.75$\pm$0.25 & 0.86$\pm$0.11 & 0.81$\pm$0.17$^{\ }$
& 33.9 / 0.12 & 0.7 / 0.03 & 12.2 / 0.06
& 5.5 / 0.04 & 1.4 / 0.03 & 3.6 / 0.03& 40\%\\

\bottomrule
\end{tabular}%
}

\label{tab:methods_rows_fullcost}
\end{table*}
\noindent\textbf{1) 2D Retinal Fundus Disc Segmentation.}
We used five real clinical sites following~\cite{jiang2023fair}: Drishti-GS1 (site \textit{A})~\cite{drishti}, RIGA (sites \textit{B--D})~\cite{riga}, and REFUGE (site \textit{E})~\cite{refuge}. The datasets contain color fundus images from healthy and glaucoma patients with expert optic disc annotations. Images were resized to $256 \times 256$ and normalized to $[0,1]$.

\noindent
\textbf{2) 3D Brain Tumor Segmentation.}
We used the BraTS Post-Treatment Glioma Challenge dataset~\cite{brats}, consisting of 1,942 multi-parametric MRI acquired from five clinical centers: UC San Francisco (site \textit{P}), Duke University (site \textit{Q}), UC San Diego (site \textit{R}), University of Missouri (site \textit{S}), and Indiana University (site \textit{T}). Of the 1,942 scans 1,621 scans were released with ground truth for training, which we used in our analysis. Each case includes four MRI modalities: native T1, post-contrast T1-weighted, T2-weighted, and T2-FLAIR. All scans were preprocessed using co-registration to a common template, resampling to $1\text{mm}^3$ isotropic resolution, and skull stripping using the BraTS pipeline~\cite{10.3389/fnins.2020.00125}. The segmentation targets were Enhancing Tumor (ET), Tumor Core (TC), and Whole Tumor (WT). 

\paragraph{\textbf{Metrics}}
Performance was measured using the Dice Similarity Coefficient (DSC), calculated case-wise, and averaged per site. Statistical significance was assessed using the paired Wilcoxon signed-rank test ($p<0.05$). 

The compute cost was measured in Giga Multiply-Accumulate operations (GMAC) for a common reference batch size of 10. GMAC is a billion multiply and accumulate operations ($a \times b + c$), a standard proxy for deep learning hardware computation. We report GMAC as a hardware-agnostic proxy for computational cost, allowing a consistent comparison of model efficiency across heterogeneous clinical and edge-device settings where runtime measurements depend strongly on hardware, software, and system load. Training GMAC includes forward, backward, and weight update computation, whereas inference GMAC includes forward computation only.

In addition, we also estimated peak VRAM usage. Specifically, we approximated the memory demand as the sum of model weights, gradients, optimizer states, activation tensors, and temporary workspace buffer of 15\% with \texttt{torch.float16} datatype. This provides an estimate of memory usage without runtime profiling.

The communication cost was quantified as the average number of parameters exchanged per FL round and reported as a percentage relative to \textit{FedAvg}. For \textit{Fed-ADApt} and \textit{DepthFL}, each client uploads and downloads only the parameters corresponding to its instantiated depth, while the server maintains a unified hierarchical global model.

\paragraph{\textbf{Implementation Details}}
All models were trained for 300 FL rounds using \textit{AdamW}~\cite{loshchilov2017decoupled} optimizer and Dice loss function with default \textit{Lightning v2.6.0} and \textit{MONAI v1.5.1}~\cite{MONAI} settings. A \textit{cosine annealing} learning rate scheduler was used. We extended \textit{MONAI UNet} to support multi-depth architectures (Sec.~\ref{sec:adapt_unet}, code available\footnote{\begingroup\scriptsize\url{https://github.com/Pediatric-Accelerated-Intelligence-Lab/MultiDepthUNet}\endgroup}). Data augmentation included random intensity scaling ($\pm0.1$), intensity shifting ($\pm0.1$), and 50\% random axis flips. For 2D retinal disc segmentation, the models were trained with batch size 96 and $lr = 1e^{-3}$. For 3D brain tumor segmentation, we used $160^3$ patches at 1mm$^3$ resolution, batch size 4 and $lr = 1e^{-4}$. 

\noindent\textit{\textbf{Baselines:}}
We compared with \textit{FedAvg} (homogeneous full-capacity FL) and independently trained local \textit{UNets}. In the local setting, each site selects a depth $l \le C_s^{\mathrm{train}}$, providing a lower bound for no FL collaboration. We further evaluate transferring a high resource model to constrained sites to assess whether collaborative depth-adaptive training offers benefits beyond naive model transfer.

We focus on depth-adaptive heterogeneous baselines because depth naturally corresponds to hierarchical representations in \textit{UNet}. In contrast, width-adaptive methods, such as \textit{HeteroFL}\cite{diao2021heterofl} prune channel dimensions across clients, which can disrupt encoder--decoder feature compatibility in skip connections unless the architecture is substantially redesigned. Therefore, \textit{DepthFL}\cite{kim2023depthfl} provides the closest heterogeneous FL baseline to evaluate depth-adaptive training under compute constraints.

\paragraph{\textbf{2D Retinal Disc Segmentation}}
Table~\ref{tab:methods_rows_fullcost} shows full-resource baselines \textit{FedAvg} and \textit{federated anytime UNet} achieved the highest Dice scores (0.85 and 0.84, respectively).  \textit{Fed-ADApt} closely followed with a Dice of 0.81, significantly outperforming \textit{DepthFL}, the high resource model, and the local model. Notably, at the lowest compute site C,  \textit{Fed-ADApt} improved Dice from 0.20 with local training and 0.45 with  \textit{DepthFL} to 0.86, indicating that multi-depth supervision and federated aggregation substantially improve worst-site performance under severe compute constraints. The high-resource model performed poorly on several 2D sites, highlighting that compute-rich single-site training does not overcome cross-site domain shift.
\begin{table*}[t]

\caption{\textbf{3D brain tumor segmentation.} Dice scores for enhancing tumor (ET), tumor core (TC), and whole tumor (WT) for various FL methods. GMAC was calculated for a batch size of 10. $^*$ indicates statistical significance ($p < 0.05$) in WT Dice compared to \textit{Fed-ADApt}.  Training strategy:
\legendbox{BlockA} No FL;
\legendbox{BlockB} full-capacity FL;
\legendbox{BlockC} compute-aware FL.}
\centering
\resizebox{\textwidth}{!}{%
\begin{tabular}{lccccccc||ccc|ccc|c}
\toprule
& \multicolumn{7}{c||}{\textbf{Segmentation}}
& \multicolumn{3}{c|}{\textbf{Training}}
& \multicolumn{3}{c|}{\textbf{Inference}}& \textbf{Comm.} \\
\textbf{Method} 
& \multicolumn{7}{c||}{\textbf{(Dice)}}
& \multicolumn{3}{c|}{\textbf{GMAC / VRAM(GB)}}
& \multicolumn{3}{c|}{\textbf{GMAC / VRAM(GB)}} & \textbf{Cost}\\

 & &Site P & Site Q & Site R & Site S & Site T
& Avg& Max & Min & Avg & Max & Min & Avg &  \\
\midrule

\rowcolor{BlockA}
& ET
& 0.81$\pm$0.13 & 0.85$\pm$0.12 & 0.71$\pm$0.22 & 0.74$\pm$0.23 & 0.33$\pm$0.14 & 0.69$\pm$0.17$^{\ }$
&&&&&&&\\

\rowcolor{BlockA}
Local Model & TC
& 0.83$\pm$0.12 & 0.87$\pm$0.10 & 0.74$\pm$0.22 & 0.77$\pm$0.21 & 0.50$\pm$0.20 & 0.74$\pm$0.17$^{\ }$
&477 / 0.99&133 / 0.80& 357 / 0.94
&159 / 0.97&44 / 0.79&119 / 0.93& -\\
\rowcolor{BlockA}
& WT
& 0.83$\pm$0.12 & 0.88$\pm$0.11 & 0.75$\pm$0.22 & 0.77$\pm$0.22 & 0.10$\pm$0.04 & 0.66$\pm$0.16$^*$
&&&&&&& \\
\tightcmidrule{2-15}

\rowcolor{BlockA}
& ET
& 0.81$\pm$0.13 & 0.73$\pm$0.22 & 0.68$\pm$0.24 & 0.56$\pm$0.29 & 0.74$\pm$0.14 & 0.71$\pm$0.20$^{\ }$
&&&&&&&\\
\rowcolor{BlockA}
High Resource Model
& TC
& 0.83$\pm$0.12 & 0.73$\pm$0.23 & 0.70$\pm$0.24 & 0.59$\pm$0.29 & 0.81$\pm$0.08 & 0.73$\pm$0.17$^{\ }$
&477 / 0.99&0 / 0.00 &96 / 0.25&159 / 0.97&159 / 0.97&159 / 0.97&-\\
\rowcolor{BlockA}
& WT
& 0.83$\pm$0.12 & 0.73$\pm$0.23 & 0.71$\pm$0.25 & 0.59$\pm$0.29 & 0.84$\pm$0.10 & 0.74$\pm$0.22$^*$
&&&&&&& \\
\rowcolor{BlockB}

& ET
& 0.80$\pm$0.14 & 0.84$\pm$0.13 & 0.72$\pm$0.23 & 0.73$\pm$0.26 & 0.78$\pm$0.12 & 0.78$\pm$0.17  &  & & &  &  &  &\\
\rowcolor{BlockB}
FedAvg
& TC
& 0.81$\pm$0.12 & 0.87$\pm$0.11 & 0.76$\pm$0.21 & 0.76$\pm$0.25 & 0.85$\pm$0.08 & 0.81$\pm$0.15
&477 / 0.99&477 / 0.99&477 / 0.99&159 / 0.97&159 / 0.97&159 / 0.97& 100\%\\
\rowcolor{BlockB}
& WT
& 0.81$\pm$0.12 & 0.87$\pm$0.11 & 0.77$\pm$0.22 & 0.77$\pm$0.26 & 0.88$\pm$0.09 & 0.82$\pm$0.16
&&&&&&& \\
\tightcmidrule{2-15}

\rowcolor{BlockB}
& ET
& 0.80$\pm$0.14 & 0.85$\pm$0.13 & 0.73$\pm$0.23 & 0.75$\pm$0.23 & 0.79$\pm$0.13 & 0.79$\pm0.17$ &  & & &  &  &&  \\
\rowcolor{BlockB}
Federated Anytime & TC
& 0.81$\pm$0.13 & 0.88$\pm$0.09 & 0.76$\pm$0.21 & 0.78$\pm$0.22 & 0.86$\pm$0.08 & 0.82$\pm$0.13
&1266 / 3.76&1266 / 3.76&1266 / 3.76&159 / 0.97&143 / 0.97&152 / 0.97& 100\%\\
\rowcolor{BlockB}
& WT
& 0.82$\pm$0.13 & 0.88$\pm$0.10 & 0.77$\pm$0.22 & 0.79$\pm$0.23 & 0.89$\pm$0.09 & 0.83$\pm$0.12
&&&&&&& \\

\rowcolor{BlockC}

& ET
& 0.80$\pm$0.13 & 0.82$\pm$0.15 & 0.68$\pm$0.23 & 0.70$\pm$0.25 & 0.77$\pm$0.11 & 0.75$\pm$0.16&  & & &  &  &  &\\
\rowcolor{BlockC}
DepthFL & TC
& 0.82$\pm$0.12 & 0.85$\pm$0.13 & 0.72$\pm$0.22 & 0.73$\pm$0.24 & 0.84$\pm$0.07& 0.79$\pm$0.15
&477 / 0.99&133 / 0.80&357 / 0.94&159 / 0.97&159 / 0.97&159 / 0.97&75\% \\
\rowcolor{BlockC}
& WT
& 0.82$\pm$0.12 & 0.85$\pm$0.13 & 0.73$\pm$0.23 & 0.74$\pm$0.24 & 0.87$\pm$0.08 & 0.80$\pm$0.16
&&&&&&& \\
\tightcmidrule{2-15}

\rowcolor{BlockC}

& ET
& 0.79$\pm$0.14 & 0.85$\pm$0.11 & 0.71$\pm$0.22 & 0.74$\pm$0.24 & 0.79$\pm$0.10 & 0.77$\pm$0.16
&&&&&&& \\
\rowcolor{BlockC}
\textbf{Fed-ADApt (Ours)}& TC
& 0.80$\pm$0.12 & 0.87$\pm$0.09 & 0.74$\pm$0.20 & 0.77$\pm$0.22 & 0.85$\pm$0.13 & 0.81$\pm$0.15& 1266 / 3.76&133 / 0.80&720 / 2.43& 144 / 0.97 & 124 / 0.96 & 128 / 0.96 &75\%\\

\rowcolor{BlockC}
& WT
& 0.81$\pm$0.13 & 0.87$\pm$0.10 & 0.75$\pm$0.21 & 0.77$\pm$0.22 & 0.88$\pm$0.09 & 0.82$\pm$0.15&
&&&&&& \\
\bottomrule
\end{tabular}
}
\label{tab:brain_results}
\end{table*}

In terms of compute cost, \textit{FedAvg} required a fixed 16.5 GMAC for training and 5.5 GMAC for inference at all sites, while \textit{federated anytime UNet} incurred the highest training cost (33.9 GMAC). This shows that adaptive inference alone is insufficient for low-resource FL, since federated anytime still requires every client to train the full multi-depth model. In contrast, \textit{Fed-ADApt} adapted to site-specific resources, with training costs ranging from 0.7 to 33.9 GMAC (12.2 GMAC on average) and an average inference cost of 3.6 GMAC (3.6, 1.4, 3.6, 3.6, and 5.5 GMAC for Sites A-E, respectively). The selected inference depths are not required to equal the
training depths because inference is forward-only and substantially
cheaper than training; therefore, constrained sites may use deeper
prediction paths at deployment than they can optimize locally.

The estimated VRAM requirements show a similar resource-adaptive trend: \textit{Fed-ADApt} reduces the average training memory demand to 0.06 GB compared to 0.12 GB for \textit{federated anytime UNet}, while constrained clients require as little as 0.03 GB.

\begin{figure}[h!]
\centering
\includegraphics[width=\columnwidth]{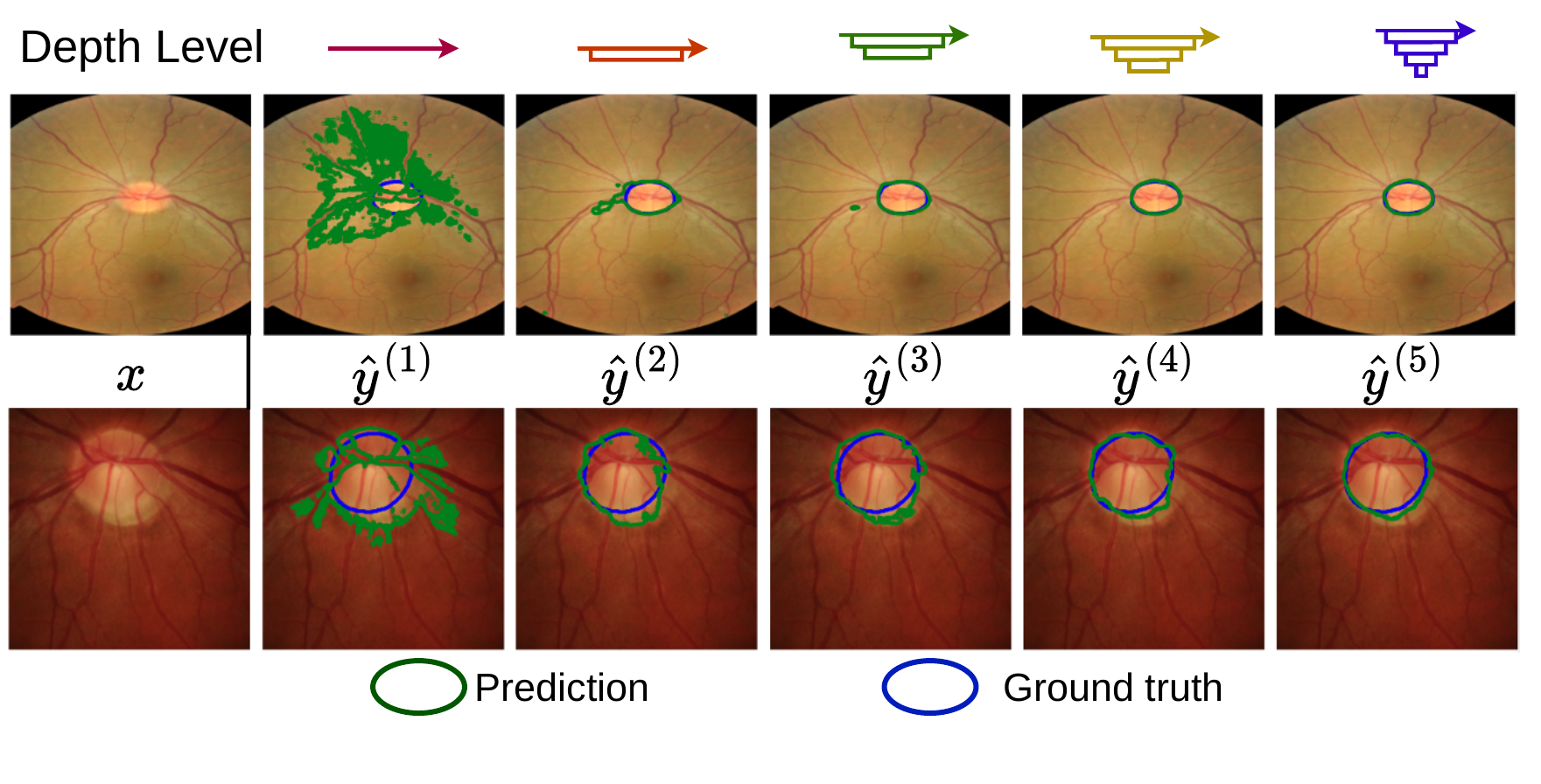}
\caption{\textbf{Qualitative effect of depth-adaptive inference.} Retinal fundus predictions from Fed-ADApt at increasing depth levels. Shallower prediction paths provide coarse optic disc localization with lower computation, while deeper paths progressively refine the segmentation boundary and better align with the expert annotation. Green contours indicate predictions and blue contours indicate ground truth.}\label{fig:fedadapt}
\end{figure}

This visual pattern is consistent with Fig.~\ref{fig:fedadapt}, where increasing depth improves boundary delineation while retaining the option of shallower inference under constrained compute budgets.

Despite only a 4.7\% reduction in average Dice compared to \textit{FedAvg}, \textit{Fed-ADApt} substantially reduced compute and communication costs. In particular, at the smallest site Site~C, it reduced the training cost by 96\% compared to \textit{FedAvg} (16.5 to 0.7 GMAC) and by 98\% compared to \textit{federated anytime UNet} (33.9 to 0.7 GMAC). Furthermore, in all sites, \textit{Fed-ADApt} reduced the inference cost by 34.5\% compared to \textit{FedAvg} at the smallest site, demonstrating a favorable trade-off between accuracy and efficiency.

%

\begin{table}[h!]
\centering
\caption{\textbf{Sensitivity of Fed-ADApt to federation compute budget.}
Site-level $C_s^{\mathrm{train}}$ and Dice scores are shown for low-, medium-, and high-budget federations. Increasing the federation budget improves mean Dice.}
\label{tab:compute_dice}
\resizebox{0.95\columnwidth}{!}{%
\begin{tabular}{lcc||cc||cc}
\toprule
\multirow{2}{*}{} 
& \multicolumn{2}{c||}{\textbf{Low budget}} 
& \multicolumn{2}{c||}{\textbf{Medium budget}} 
& \multicolumn{2}{c}{\textbf{High budget}} \\
& $C_s^{\mathrm{train}}$ & Dice 
& $C_s^{\mathrm{train}}$ & Dice 
& $C_s^{\mathrm{train}}$ & Dice \\
\midrule
Site A   & 1   & 0.86 & 3   & 0.86 & 4   & 0.91 \\
Site B   & 2   & 0.68 & 3   & 0.78 & 5   & 0.73 \\
Site C   & 1   & 0.86 & 1   & 0.87 & 1   & 0.89 \\
Site D   & 4   & 0.75 & 4   & 0.78 & 4   & 0.81 \\
Site E   & 5   & 0.86 & 5   & 0.84 & 5   & 0.85 \\
\midrule
Avg & 2.6 & 0.80 & 3.2 & 0.82 & 3.8 & 0.84 \\
\bottomrule
\end{tabular}
}
\end{table}
Table~\ref{tab:compute_dice} summarizes the impact of the compute budget $C_s^{\mathrm{train}}$ at $s$ on segmentation. Increasing the compute budget $C_s^{\mathrm{train}}$ independent of the site size, while keeping the highest and lowest resource sites fixed, led to a steady improvement in Dice from 0.80 to 0.82 and 0.84, respectively. This indicates that \textit{Fed-ADApt} does not merely reduce computation, but can also exploit additional client capacity when available. At the site level, Site \textit{C} showed consistent gains with increasing compute: 0.86, 0.87, and 0.89 for low, medium, and high settings, while Site \textit{E} remained largely stable across settings.

\paragraph{\textbf{3D Brain Tumor Segmentation}}
Table~\ref{tab:brain_results} shows the full-resource \textit{FedAvg} and \textit{federated anytime} achieved the highest WT Dice scores (0.82 and 0.83, respectively). \textit{Fed-ADApt} matched \textit{FedAvg} with a WT Dice of 0.82, while significantly outperforming the high-resource and the local models. This demonstrates that, in volumetric segmentation, Fed-ADApt can preserve full-resource FL performance while reducing resource requirements. For compute costs, \textit{FedAvg} required a fixed 477 GMAC for training and 159 GMAC for inference across all sites. In contrast, \textit{Fed-ADApt} adapted to heterogeneous site resources, with training costs ranging from 133 to 1266 GMAC and an average inference cost of 128 GMAC (144, 124, 128, 128, and 124 GMAC for Sites \textit{P}--\textit{T}, respectively). \textit{Fed-ADApt} also reduces the average estimated training VRAM to 2.43 GB, compared with 3.76 GB for \textit{federated anytime UNet}, while the most constrained client requires only 0.80 GB.

While \textit{Fed-ADApt} matched the WT Dice performance of \textit{FedAvg}, at all sites, \textit{Fed-ADApt} decreased the inference cost by 22\% at the smallest site and communication cost to 75\% of \textit{FedAvg}, i.e., a 25\% reduction. At the smallest site, Site \textit{T}, it reduced the training cost by 72\% compared to \textit{FedAvg} (477 to 133 GMAC). The higher inference GMAC at some low-training-budget sites reflects validation-based selection of a deeper deployment path; this remains feasible because inference is forward-only and substantially cheaper than training.

\section{Discussion}

We propose \textit{Fed-ADApt} to address a key limitation of heterogeneous FL for medical image segmentation: supporting variability in both training and inference compute budgets and reducing training communications for the \textit{UNet} architecture. Across both 2D and 3D tasks, FL outperformed non-collaborative training. Local models at low-resource sites performed poorly because shallow networks lacked sufficient capacity to learn discriminative features. In contrast, the high resource model trained at the largest site generalized poorly due to cross-site variability, especially in retinal segmentation, whereas the brain tumor dataset showed fewer shifts due to high-quality preprocessing. 

Our \textit{federated anytime UNet} leverages implicit deep supervision that shapes early-layer representations via additional gradient propagation. 
This implicit deep supervision favors learning features that remain predictive even under truncated inference. 
In contrast, \textit{DepthFL} supervises only the maximum instantiated depth at each site, meaning shallower layers at low compute sites are optimized without consistent guidance from high compute sites. 
Building on inference adaptivity, \textit{Fed-ADApt} extends compute awareness to the training phase and communication cost, enabling smaller sites that would otherwise be excluded to contribute and benefit from global updates. Importantly, this resource adaptivity extends beyond arithmetic computation. By restricting the instantiated network depth, \textit{Fed-ADApt} also reduces the estimated memory required for parameters, activations, gradients, and optimizer states, addressing a second practical barrier to training at resource-constrained sites.


Beyond accuracy, \textit{Fed-ADApt} effectively utilizes additional compute resources when available. Increasing the federation's compute budget improved Dice performance, indicating that greater depth enhances representation and segmentation quality. However, gains plateaued at some sites (e.g., 2D task at Site~\textit{E}), suggesting diminishing returns beyond moderate capacity increase. This saturation effect suggests that, using \textit{Fed-ADApt}, compute resources can be strategically allocated rather than uniformly increased. \textit{Fed-ADApt} maintains competitive worst-site performance compared to full-resource FL while reducing compute and communication cost. 

In general, these findings position \textit{Fed-ADApt}  as a practical and equitable framework for deployable FL segmentation under real-world constraints, including resource-constrained clinical and edge-enabled imaging environments. The reductions in estimated VRAM are particularly relevant for edge and IoMT deployment, where available accelerator memory can be as limiting as computational throughput. Because these values are analytically estimated, direct evaluation on representative IoMT hardware remains necessary to establish realized latency, memory, and energy benefits. By allowing constrained clients to train shallow subnetworks and select shallower prediction paths at inference, \textit{Fed-ADApt}  provides a step toward federated segmentation across heterogeneous infrastructure, from high-resource hospitals to edge-connected imaging devices.

Nevertheless, \textit{Fed-ADApt} assumes predefined compute budgets rather than dynamically estimating them, motivating future work on automated budget estimation, as well as deployment on real IoMT hardware and extensions to hierarchical UNet-family or transformer-hybrid segmentation architectures such as \textit{UNETR} and \textit{SwinUNETR}. In addition, an uncertainty-based depth selection strategy could be studied instead of the validation score.

Several limitations remain. First, Fed-ADApt currently assumes predefined compute budgets instead of dynamically estimating client hardware constraints. Second, the evaluation was performed using simulated compute assignments rather than deployment on real IoMT or edge hardware. Third, inference depth was selected using validation performance; future work could explore uncertainty-aware or latency-aware depth selection. Finally, extending Fed-ADApt to other hierarchical segmentation architectures, including UNETR, SwinUNETR, and transformer-hybrid UNet variants, would further test the generality of the framework.

\section{Conclusion}

\textit{Fed-ADApt}  introduces a \textit{UNet}-compatible depth-adaptive federated learning framework for resource-aware medical image segmentation. Unlike direct depth truncation methods developed for classification models, \textit{Fed-ADApt} defines depth-specific full-resolution prediction paths, preserves encoder–decoder skip connections, and aggregates parameters hierarchically according to each client’s training budget. This allows low-compute sites to participate in federated training without requiring full-capacity model optimization.

At the same time, \textit{Fed-ADApt}  produces a global multi-depth model that supports deployment-time depth selection, enabling sites to balance inference cost and segmentation accuracy without retraining. Across 2D retinal disc and 3D brain tumor segmentation, Fed-ADApt preserved competitive segmentation performance while substantially reducing resource requirements for constrained clients and lowering inference and communication costs relative to full-model FL. 

Overall, \textit{Fed-ADApt} provides a practical step toward equitable federated medical image segmentation across heterogeneous clinical institutions and edge-enabled imaging environments.

\section*{Acknowledgments}
\label{sec:acknowledgments}
This work was partially supported by funds received from the \textit{Gilbert Family Foundation}. 

\bibliographystyle{IEEEbib}
\bibliography{refs}

@misc{cost,
title={What’s the backward-forward FLOP ratio for neural networks?},
author={Marius Hobbhahn and Jaime Sevilla},
year={2021},
url={https://epoch.ai/publications/backward-forward-FLOP-ratio},
note={Accessed: 2026-07-01}}

@inproceedings{kaur2023federated,
  title={Federated learning in IoT: A survey from a resource-constrained perspective},
  author={Kaur, Ishmeet and Jadhav, Adwaita Janardhan},
  booktitle={2023 International Conference on Artificial Intelligence Robotics, Signal and Image Processing (AIRoSIP)},
  pages={376--381},
  year={2023},
  organization={IEEE}
}

@article{kagita2022review,
  title={A review on security and privacy of internet of medical things},
  author={Kagita, Mohan Krishna and Thilakarathne, Navod and Gadekallu, Thippa Reddy and Maddikunta, Praveen Kumar Reddy},
  journal={Intelligent internet of things for healthcare and industry},
  pages={171--187},
  year={2022},
  publisher={Springer}
}

@article{deoni2022development,
  title={Development of a mobile low-field MRI scanner},
  author={Deoni, Sean CL and Medeiros, Paul and Deoni, Alexandra T and Burton, Phoebe and Beauchemin, Jennifer and D’Sa, Viren and Boskamp, Eddy and By, Samantha and McNulty, Chris and Mileski, William and others},
  journal={Scientific reports},
  volume={12},
  number={1},
  pages={5690},
  year={2022},
  publisher={Nature Publishing Group UK London}
}

@article{chen2025acceptability,
  title={Acceptability, applicability, and cost-utility of artificial-intelligence-powered low-cost portable fundus camera for diabetic retinopathy screening in primary health care settings},
  author={Chen, Yanxian and Song, Fan and Zhao, Ziwei and Wang, Yueye and To, Elaine and Liu, Yanjun and Shi, Danli and Chen, Xiaolan and Xu, Liya and Shang, Xianwen and others},
  journal={Diabetes research and clinical practice},
  volume={223},
  pages={112161},
  year={2025},
  publisher={Elsevier}
}

@inproceedings{kim2023depthfl,
  title={Depthfl: Depthwise federated learning for heterogeneous clients},
  author={Kim, Minjae and Yu, Sangyoon and Kim, Suhyun and Moon, Soo-Mook},
  booktitle={The Eleventh International Conference on Learning Representations, ICLR 2023},
  year={2023}
}

@article{liu2022federated,
  title={Federated learning for inference at anytime and anywhere},
  author={Liu, Zicheng and Li, Da and Fernandez-Marques, Javier and Laskaridis, Stefanos and Gao, Yan and Li, Stan Z and Hu, Shell Xu and Hospedales, Timothy and others},
  journal={arXiv preprint arXiv:2212.04084},
  year={2022}
}

@ARTICLE{10.3389/fnins.2020.00125,
    
AUTHOR={Kofler, Florian  and Berger, Christoph  and Waldmannstetter, Diana  and Lipkova, Jana  and Ezhov, Ivan  and Tetteh, Giles  and Kirschke, Jan  and Zimmer, Claus  and Wiestler, Benedikt  and Menze, Bjoern H. },
           
TITLE={BraTS Toolkit: Translating BraTS Brain Tumor Segmentation Algorithms Into Clinical and Scientific Practice},
          
JOURNAL={Frontiers in Neuroscience},
          
VOLUME={Volume 14 - 2020},
  
YEAR={2020},
  
URL={https://www.frontiersin.org/journals/neuroscience/articles/10.3389/fnins.2020.00125},
  
DOI={10.3389/fnins.2020.00125},
  
ISSN={1662-453X}}

@inproceedings{zhang2023memory,
  title={Memory-adaptive Depth-wise Heterogenous Federated Learning},
  author={Zhang, Kai and Dai, Yutong and Wang, Hongyi and Xing, Eric and Chen, Xun and Sun, Lichao},
  booktitle={Federated Learning Systems (FLSys) Workshop@ MLSys 2023}
}

@inproceedings{diao2021heterofl,
  title={HETEROFL: COMPUTATION AND COMMUNICATION EFFICIENT FEDERATED LEARNING FOR HETEROGENEOUS CLIENTS},
  author={Diao, Enmao and Ding, Jie and Tarokh, Vahid},
  booktitle={The Ninth International Conference on Learning Representations, ICLR 2021},
  year={2021}
}

@inproceedings{lee2024recurrent,
  title={Recurrent Early Exits for Federated Learning with Heterogeneous Clients},
  author={Lee, Royson and Fernandez-Marques, Javier and Hu, Shell Xu and Li, Da and Laskaridis, Stefanos and Dudziak, {\L}ukasz and Hospedales, Timothy and Husz{\'a}r, Ferenc and Lane, Nicholas Donald},
  booktitle={International Conference on Machine Learning},
  pages={26568--26588},
  year={2024},
  organization={PMLR}
}

@inproceedings{unet,
  title={U-net: Convolutional networks for biomedical image segmentation},
  author={Ronneberger, Olaf and Fischer, Philipp and Brox, Thomas},
  booktitle={Medical image computing and computer-assisted intervention--MICCAI 2015: 18th international conference, Munich, Germany, October 5-9, 2015, proceedings, part III 18},
  pages={234--241},
  year={2015},
  organization={Springer}
}

@inproceedings{fedavg,
  title={Communication-efficient learning of deep networks from decentralized data},
  author={McMahan, Brendan and Moore, Eider and Ramage, Daniel and Hampson, Seth and y Arcas, Blaise Aguera},
  booktitle={Artificial intelligence and statistics},
  pages={1273--1282},
  year={2017},
  organization={PMLR}
}

@inproceedings{qi2022contrastive,
  title={Contrastive re-localization and history distillation in federated cmr segmentation},
  author={Qi, Xiaoming and Yang, Guanyu and He, Yuting and Liu, Wangyan and Islam, Ali and Li, Shuo},
  booktitle={International conference on medical image computing and computer-assisted intervention},
  pages={256--265},
  year={2022},
  organization={Springer}
}

@software{MONAI,
author = {{MONAI Consortium}},
license = {Apache-2.0},
month = {9},
title = {{MONAI: Medical Open Network for AI}},
url = {https://github.com/Project-MONAI/MONAI},
version = {1.0.0},
year = {2022}
}

@article{wu2022communication,
  title={Communication-efficient federated learning via knowledge distillation},
  author={Wu, Chuhan and Wu, Fangzhao and Lyu, Lingjuan and Huang, Yongfeng and Xie, Xing},
  journal={Nature communications},
  volume={13},
  number={1},
  pages={2032},
  year={2022},
  publisher={Nature Publishing Group UK London}
}

@article{SUN2025127973,
title = {FedAgent: Federated Learning on Non-IID Data via Reinforcement Learning and Knowledge Distillation},
journal = {Expert Systems with Applications},
volume = {285},
pages = {127973},
year = {2025},
issn = {0957-4174},
doi = {https://doi.org/10.1016/j.eswa.2025.127973},
url = {https://www.sciencedirect.com/science/article/pii/S0957417425015957},
author = {Bingli Sun and Xiao Song and Yuchun Tu and Ming Liu}
}

@article{brats,
  title={The 2024 brain tumor segmentation (brats) challenge: Glioma segmentation on post-treatment mri},
  author={de Verdier, Maria Correia and Saluja, Rachit and Gagnon, Louis and LaBella, Dominic and Baid, Ujjwall and Tahon, Nourel Hoda and Foltyn-Dumitru, Martha and Zhang, Jikai and Alafif, Maram and Baig, Saif and others},
  journal={arXiv preprint arXiv:2405.18368},
  year={2024}
}

@article{loshchilov2017decoupled,
  title={Decoupled weight decay regularization},
  author={Loshchilov, Ilya and Hutter, Frank},
  journal={arXiv preprint arXiv:1711.05101},
  year={2017}
}

@inproceedings{riga,
  title={Retinal fundus images for glaucoma analysis: the RIGA dataset},
  author={Almazroa, Ahmed and Alodhayb, Sami and Osman, Essameldin and Ramadan, Eslam and Hummadi, Mohammed and Dlaim, Mohammed and Alkatee, Muhannad and Raahemifar, Kaamran and Lakshminarayanan, Vasudevan},
  booktitle={Medical Imaging 2018: Imaging Informatics for Healthcare, Research, and Applications},
  volume={10579},
  pages={55--62},
  year={2018},
  organization={SPIE}
}

@ARTICLE{10004567,
  author={Wicaksana, Jeffry and Yan, Zengqiang and Zhang, Dong and Huang, Xijie and Wu, Huimin and Yang, Xin and Cheng, Kwang-Ting},
  journal={IEEE Transactions on Medical Imaging}, 
  title={FedMix: Mixed Supervised Federated Learning for Medical Image Segmentation}, 
  year={2023},
  volume={42},
  number={7},
  pages={1955-1968},
  doi={10.1109/TMI.2022.3233405}}

@ARTICLE{3235757,
  author={Zhu, Meilu and Chen, Zhen and Yuan, Yixuan},
  journal={IEEE Transactions on Medical Imaging}, 
  title={FedDM: Federated Weakly Supervised Segmentation via Annotation Calibration and Gradient De-conflicting}, 
  year={2023},
  doi={10.1109/TMI.2023.3235757}}

@inproceedings{jiang2023fair,
  title={Fair federated medical image segmentation via client contribution estimation},
  author={Jiang, Meirui and Roth, Holger R and Li, Wenqi and Yang, Dong and Zhao, Can and Nath, Vishwesh and Xu, Daguang and Dou, Qi and Xu, Ziyue},
  booktitle={Proceedings of the IEEE/CVF Conference on Computer Vision and Pattern Recognition},
  pages={16302--16311},
  year={2023}
}

@article{GUAN2024110424,
title = {Federated learning for medical image analysis: A survey},
journal = {Pattern Recognition},
volume = {151},
pages = {110424},
year = {2024},
issn = {0031-3203},
doi = {https://doi.org/10.1016/j.patcog.2024.110424},
url = {https://www.sciencedirect.com/science/article/pii/S0031320324001754},
author = {Hao Guan and Pew-Thian Yap and Andrea Bozoki and Mingxia Liu}
}

@article{litany2022federated,
  title={Federated learning with heterogeneous architectures using graph hypernetworks},
  author={Litany, Or and Maron, Haggai and Acuna, David and Kautz, Jan and Chechik, Gal and Fidler, Sanja},
  journal={arXiv preprint arXiv:2201.08459},
  year={2022}
}

@article{10.1145/3625558,
author = {Ye, Mang and Fang, Xiuwen and Du, Bo and Yuen, Pong C. and Tao, Dacheng},
title = {Heterogeneous Federated Learning: State-of-the-art and Research Challenges},
year = {2023},
issue_date = {March 2024},
publisher = {Association for Computing Machinery},
address = {New York, NY, USA},
volume = {56},
number = {3},
issn = {0360-0300},
url = {https://doi.org/10.1145/3625558},
doi = {10.1145/3625558},
journal = {ACM Comput. Surv.},
month = oct,
articleno = {79},
numpages = {44}
}

@INPROCEEDINGS{10825769,
  author={Fan, Boyu and Jiang, Siyang and Su, Xiang and Tarkoma, Sasu and Hui, Pan},
  booktitle={2024 IEEE International Conference on Big Data (BigData)}, 
  title={A Survey on Model-heterogeneous Federated Learning: Problems, Methods, and Prospects}, 
  year={2024},
  volume={},
  number={},
  pages={7725-7734},
  doi={10.1109/BigData62323.2024.10825769}}

@article{refuge,
  title={Refuge challenge: A unified framework for evaluating automated methods for glaucoma assessment from fundus photographs},
  author={Orlando, Jos{\'e} Ignacio and Fu, Huazhu and Breda, Jo{\~a}o Barbosa and Van Keer, Karel and Bathula, Deepti R and Diaz-Pinto, Andr{\'e}s and Fang, Ruogu and Heng, Pheng-Ann and Kim, Jeyoung and Lee, JoonHo and others},
  journal={Medical image analysis},
  volume={59},
  pages={101570},
  year={2020},
  publisher={Elsevier}
}

@inproceedings{drishti,
  author={Sivaswamy, Jayanthi and Krishnadas, S. R. and Datt Joshi, Gopal and Jain, Madhulika and Syed Tabish, A. Ujjwaft},
  booktitle={2014 IEEE 11th International Symposium on Biomedical Imaging (ISBI)}, 
  title={Drishti-GS: Retinal image dataset for optic nerve head(ONH) segmentation}, 
  year={2014},
  volume={},
  number={},
  pages={53-56},
  doi={10.1109/ISBI.2014.6867807}}

\end{document}